\documentclass[preprint,numbers]{sigplanconf}

\usepackage{amsmath}
\usepackage{booktabs} 
\usepackage{graphicx} 
\usepackage{listings} 
\usepackage[utf8]{inputenc}
\usepackage{url}
\usepackage[colorinlistoftodos,prependcaption,textsize=tiny]{todonotes}
\usepackage{fancyvrb}

\DefineShortVerb{\|}

\DefineVerbatimEnvironment%
  {Verb}{Verbatim}
  {framesep=2mm, rulecolor=\color{ForestGreen}}

\DefineVerbatimEnvironment%
  {VerbWithNumbers}{Verbatim}
  {fontsize=\small, framesep=2mm, rulecolor=\color{ForestGreen}, 
   numbers=none, numbersep=3pt, stepnumber=1}

\newcommand{\system}{AI Programmer}

\begin{document}

\setlength{\pdfpageheight}{\paperheight}
\setlength{\pdfpagewidth}{\paperwidth}

\conferenceinfo{MAPL'17}{June 18, 2017, City, ST, Country}
\copyrightyear{2017}
\copyrightdata{978-1-nnnn-nnnn-n/yy/mm}\reprintprice{\$15.00}
\copyrightdoi{nnnnnnn.nnnnnnn}


\setcopyright{acmlicensed}  



\title{AI Programmer: Autonomously Creating Software Programs Using Genetic Algorithms}


\authorinfo{Kory Becker}
           {kbecker@primaryobjects.com}
           {Bloomberg LP}
\authorinfo{Justin Gottschlich}
           {justin.gottschlich@intel.com}
           {Intel Labs}

\maketitle

\begin{abstract}



In this paper, we present the first-of-its-kind machine learning (ML) system, called AI Programmer, that can automatically generate full software programs requiring only minimal human guidance. At its core, AI Programmer uses genetic algorithms (GA) coupled with a tightly constrained programming language that minimizes the overhead of its ML search space. Part of AI Programmer's novelty stems from \emph{(i)} its unique system design, including an embedded, hand-crafted interpreter for efficiency and security and \emph{(ii)} its augmentation of GAs to include instruction-gene randomization bindings and programming language-specific genome construction and elimination techniques. We provide a detailed examination of AI Programmer's system design, several examples detailing how the system works, and experimental data demonstrating its software generation capabilities and performance using only mainstream CPUs.

\end{abstract}




\keywords
Genetic algorithm, program synthesis, genetic programming, evolutionary computation, artificial intelligence, machine learning, programming languages, code generation and optimization

\section{Introduction}
Since the invention of the computer, having the ability to correctly and efficiently develop software programs has been a principle challenge~\cite{cormen:2001:algo}. To help address this, countless breakthroughs have been made in the field of software development. Some of these include safety and flexibility advances in static, dynamic, and gradual type systems~\cite{cardelli:1985:type, siek:2006:gradual}; simplification, safety, and robustness advances using automatic memory management and garbage collection systems~\cite{dijkstra:1978:gc, jones:1996:agc}; generality and specificity progress in both general-purpose and domain specific languages~\cite{scott:2009:plp,sujeeth:2014:delite}; and, of course, a plethora of tools aimed at assisting programmers in nearly every way~\cite{graham:1982:profiler, gottschlich:2012:vtm, patil:2010:pinplay}.

Yet, simultaneous advances in hardware innovation have occurred with similar frequency, such as increasingly performant general purpose multi-core CPUs with advanced hardware extensions~\cite{mckeen:2016:sgx, yoo:2013:pei}, low power system-on-chip (SoC) edge compute devices~\cite{keating:2007:lpm}, high-performance pluggable coprocessors with near supercomputing performance of yesteryear~\cite{jeffers:2013:ixp}, wide data-parallel graphics processing units (GPUs)~\cite{nguyen:2007:gg}, and application specific integrated circuits (ASICs) for deep neural networks and computer vision~\cite{abadi:2016:tpu, stein:2005:mobileye}, to name a few. 

While such hardware advances continue to broaden and deepen the space of what is computationally tractable, they have the fracturing side-effect of complicating and exacerbating the tension between the ease of developing software and the ability for humans to write maximally efficient code. In this paper we explore an alternative approach to traditional human-driven software development; one that autonomously creates software programs using genetic algorithms (GAs) requiring only minimal human guidance.

\subsection{The Evolution of Programming Languages}

Over the last several decades, programming languages (PLs) have followed a steady path of providing higher-level programming abstractions, aimed at reducing the challenge of human-driven software development~\cite{popularPLs}. To this end, PLs, in general, have proliferated toward a design goal of simplifying human use. Although this trend is natural in an era where humans perform the majority of software development, as we will show, it is suboptimal in an environment where programming is performed predominantly by machines.

The ability for computers to automatically create their own software programs has been a long-standing goal of artificial intelligence~\cite{russell:2003:aim}. By largely removing humans from the time-intensive and error-prone process of software development and replacing them with artificial intelligence, computer software has the potential to be generated in a more streamlined, correct, and optimized fashion~\cite{lamport:1977:correct, dolan:2002:opti}.




This paper makes the following technical contributions:

\begin{enumerate}
\item We present \system\ the first-of-its-kind software generation framework, which constructs programs using genetic algorithms with novel enhancements coupled with a minimalistic programming language.
\item We present several critical observations, including an embedded interpreter and simulator solution, for security and optimization of ML-generated software.
\item We provide empirical results demonstrating the efficacy and efficiency of \system\ across several of its fully generated software programs on commodity hardware.
\end{enumerate}


\section{\label{sect:background}Background}

In this section we provide a brief synopsis of the challenges in using traditional programming languages for machine-based program generation. We also provide a brief introduction into genetic algorithms, the ML technique used by \system.

\subsection{Programming Language Density}

Most of today's programming languages were designed for human use~\cite{scott:2000:plp}. We refer to such languages as \emph{human-intended PLs} (HIPLs). Although HIPLs are useful when humans perform the majority of programming and debugging, their design is usually counter to what is needed and appropriate for ML-based PLs (MLPLs). 

HIPLs often introduce unnecessary complexity and overhead for ML program generators due, in part, to the large number of language identifiers they include. The greater the number of legal language identifiers, the greater the ML computational search space. Moreover, type systems compound the challenge of creating legal programs because variable type bindings are intentionally restrictive to protect against human error, yet provide limited value for ML program generators~\cite{pierce:2002:tpl}. For these reasons, we chose to couple AI Programmer with a non-traditional programming language that is both constrained (i.e., using only eight identifiers) and typeless. We discuss this in more detail in Section~\ref{sect:aiprogrammer}. 


Some instructions within any PL may be potentially harmful and, if used conjunction with an ML-based program generator, may cause irreversible damage. AI Programmer has specific measures in place to prevent the occurrence of such events. We discuss them in more detail in Section~\ref{sect:bfinterpret}.




\subsection{Programming with Genetic Algorithms}

A \textit{genetic algorithm} (GA) is a type of artificial intelligence, modeled after biological evolution, that begins with no knowledge of the subject, aside from an encoding of genes that represent a set of instructions or actions \cite{domingos:2015:master}. In the concept of GA-driven computer programming, a series of programming instructions are selected at random to serve as an initial chain of DNA. The complete genome is executed as a program, with the resulting fitness score calculated according to how well the program can solve a given task. This is performed with a sufficiently large population size. Those that have the best fitness are mated together to produce offspring.

Each generation of programs receive extra diversity from evolutionary techniques including roulette selection, crossover, and mutation~\cite{mitchell:1998:geneticalgo}. The process is repeated at each epoch with each child generation hopefully producing more favorable results than its parents' generation until a target solution is found. Through this process, applying GAs to computer programming automation enacts a survival of the fittest model for computer program generation~\cite{michalewicz:1994:geneticalgo}. A deeper examination of these GA principles are provided in Section~\ref{sect:aiprogrammer}.




\section{\label{sect:aiprogrammer}The Design of AI Programmer}

\begin{figure*}
\centering
\includegraphics[width=175mm]{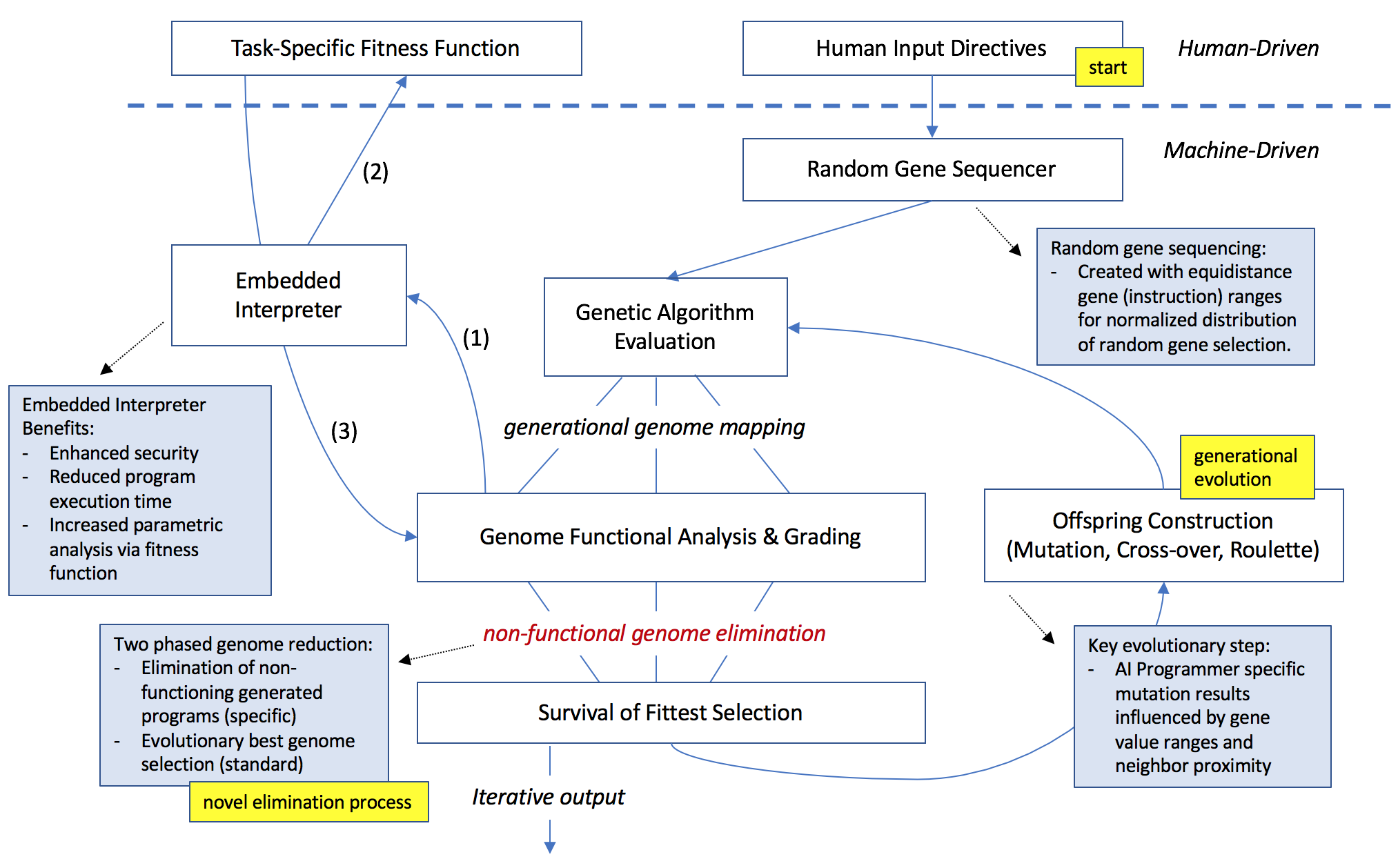}
\caption{The AI Programmer Software Architecture.}
\label{fig:bfprof_arch}
\end{figure*}

In this section we provide a high-level overview of the AI Programmer software architecture. The AI Programmer's system design is shown in Figure~\ref{fig:bfprof_arch}.

\subsection{Programming Language Selection and Challenges}

We chose a typeless programming language that contains only eight instructions to drive AI Programmer's software generation~\cite{bf:1993:wiki}. We briefly discuss the advantages of this approach and the modifications to the language that were required to integrate it into a GA solution.

\begin{table}[h!]
  \caption{AI Programmer Instruction Set and Gene Map}
  \label{tab:bfinstructionset}
  \begin{tabular}{lllll}
    \toprule
    Instr&Gene Range&Operation\\
    \midrule
    \textgreater & (0, 0.125] & Increment the pointer\\
    \textless & (0.125, 0.25] & Decrement the pointer\\
    + & (0.25, 0.375] & Increment the byte at the pointer\\
    - & (0.375, 0.5] & Decrement the byte at the pointer\\
    . & (0.5, 0.625] & Output the byte at the pointer\\
    , & (0.625, 0.75] & Input a byte and store it at the ptr\\
    {[} & (0.75, 0.875] & Jump to matching {]} if current 0\\
    {]} & (0.875, 1.0] & Jump back to matching {[} unless 0\\
  \bottomrule
\end{tabular}
\end{table}

\paragraph{Turing Completeness.}
\system's programming language, listed in  Table~\ref{tab:bfinstructionset}, is Turing complete. A \emph{Turing complete} programming language is theoretically capable of completing any (single taped Turing machine) programming task given an unlimited amount of time and memory~\cite{turing:2016:wiki}. In essence, a programming language with this characteristic is capable of implementations of a vast number of programming problems. Likewise, programs generated with AI Programmer, are theoretically capable of expressing all tasks that one might want to accomplished with computers.


\paragraph{GA Engine and Uniform Gene Distributions.}

\system's genetic algorithm engine represents each generated program's instructions as an array of floating point values, which, when considered as a unit, is its genome. Each individual location within a given genome is called a \emph{gene}. Each gene within a program's genome corresponds to a single instruction from Table~\ref{tab:bfinstructionset}. 

\system\ binds a gene value range to each of its instructions across a \emph{continuous uniform distribution} (or \emph{rectangular distribution})~\cite{uniformDist} (see Table~\ref{tab:bfinstructionset}), where each instruction's gene range is equal in size to each of the others. This was done so each instruction would have an equally random probability of being chosen at any location in a gene sequence when randomization was needed.~\footnote{We did not examine the impact of weighted ranges for different programs, but note that it may be of interest as future work.}

\paragraph{Simplified Instruction Set.}

Each of AI Programmer's instructions manipulate a memory ``tape'' of byte values, ranging from 0-255. The language works by applying increment and decrement operations to the current memory cell, while shifting the memory cell up and down the tape, as instructed by the program. The values at the current memory pointer can be input from the user or output to the terminal. Primitive looping instructions also exist (e.g., `[' and `]'), offering a complete instruction set for creating software. An example program is shown in Figure~\ref{fig:bfexample1}. 

\begin{figure}[h]
\begin{VerbWithNumbers}
+-+-+>-<[++++>+++++<+<>++]>[-[---.--[[-.++++
[+++..].]]]]
\end{VerbWithNumbers}
\caption{A generated program that outputs ``hello".}
\label{fig:bfexample1}
\end{figure}

The simplified instruction set reduces the search space in which a target program code can be found. As computational devices improve in speed, larger problem spaces can be searched. However, on less powerful devices, the search space needs to be constrained. As \system\ is intended for general purpose developers, limiting the programming instruction set to eight instructions enables the engine to execute in reasonable times on commodity hardware (see Section~\ref{sect:results}).

\subsection{Genomes and Generations}

To generate a software program using genetic algorithms, one must first create a genome. A \emph{genome} is a set of genes that are grouped together as a single unit. For AI Programmer, the genome is encoded as an array of floating point values, with fixed value ranges per unique instruction ranging between 0 and 1, as shown in the Gene Range column of Table~\ref{tab:bfinstructionset}.

Once a genome is created, it is converted to a corresponding program, executed, and the resulting program is assigned a fitness score based on the program's output. The closer a generated program comes to solving the provided task, the greater its fitness score and, the more likely it is to continue to the next evolutionary generation. At each generation epoch, AI Programmer utilizes roulette selection, along with crossover and mutation, to create child programs that contain slight random perturbations, and potentially better, genomes than their parents for solving the target task.

\begin{figure}[h]
\includegraphics[width=85mm]{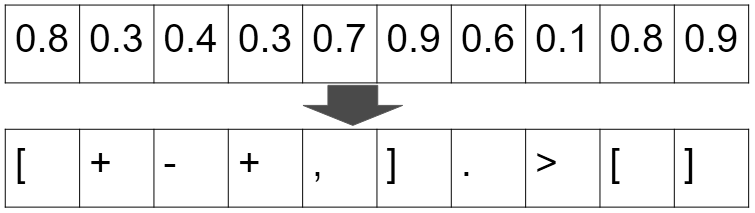}
\caption{Decoding a genome as a program.}
\label{fig:decodinggenome}
\end{figure}

\paragraph{Constructing a Genome}
Figure~\ref{fig:decodinggenome} demonstrates an example of constructing a genome from an array of floating point values. Each value range maps to a specific instruction in the programming language. Initially, these values are random (see the Random Gene Sequencer in Figure~\ref{fig:bfprof_arch}), resulting in generated programs that either won't function properly, throw errors, or simply fail \footnote{Most initial programs in the gene pool fail immediately upon being executed. Others may result in endless loops. It is due to these reasons that exception handling and maximum iteration limits are imposed on the interpreter.}. However, one or two are bound to run and execute, at a minimum, some number of valid instructions. The more successful a program is at executing, the more likely it is to continue on and produce offspring with code that achieves more successful results.

\begin{figure}[h]
\includegraphics[width=85mm]{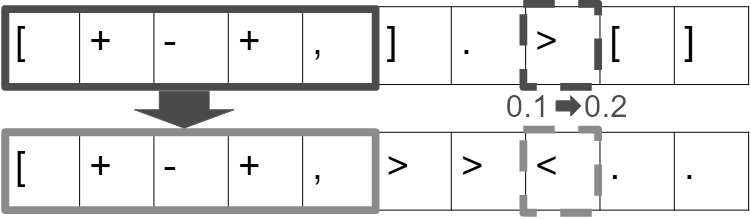}
\caption{An example of crossover and mutation. The child genome inherits the first 5 instructions from its parent. One instruction is mutated.}
\label{fig:mutation}
\end{figure}

\paragraph{Crossover and Mutation}
To create offspring, a parent genome contributes part of its genes to the child, a process called \emph{crossover}, as shown in Figure~\ref{fig:mutation}. In addition to inheriting programming instructions from its parent, each child can also experience \emph{mutation}, which is the process of adding controlled, but random perturbation, to specific genes. This results in modified behavior of the value of a particular gene, resulting in a change to the resulting programming instruction, and thus, the overall program.

Crossover copies forward potentially beneficial parts of the parent, while mutation offers differing behaviors of instruction combinations, which may or may not, end up making the child programs more successful.

\begin{figure}[h]
\includegraphics[width=85mm]{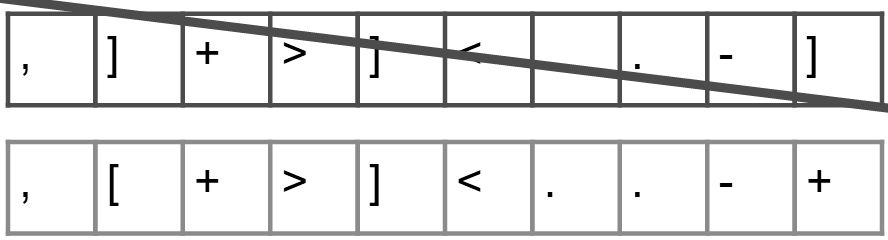}
\caption{Programs are weighted by fitness, with the most successful used for child program generation.}
\label{fig:fitness}
\end{figure}

\paragraph{Survival of the Fittest}
Executable programs are ranked according to how well they have performed. As shown in Figure~\ref{fig:fitness}, a particular program that has failed is often immediately removed from the pool of genomes. However, programs that succeed are carried forward to produce child programs.\footnote{In Figure~\ref{fig:fitness}, the bottom program is a valid running program that takes one byte for input, increments it, and then displays it twice as output.}

\subsection{\label{ssect:fitness}The Fitness Test}

To use GAs, a fitness test is needed to determine how well a generated solution performs. In the context of \system, this can involve scoring the byte-level output of the generated program, inspecting the generated program's internal state, or even analysis of intermediate state changes of the program throughout its execution. The score of the fitness test is calculated by analyzing these characteristics, and many others, and then comparing them against a user-defined target.

This concept is similar to test-driven development. When all unit tests pass, a program may be considered to be functionally correct. Likewise, a fitness test for a GA can be considered as a set of unit tests. In the case of \system, a fitness test typically contains a suite of tests for varying scenarios, which guide the genome selection, preserving only programs that evaluate well on the test suite. Further details about the construction of \system\ fitness tests are described in Section~\ref{sect:progamming}.





\subsection{\label{sect:bfinterpret}\system's Sandboxed Interpreter}






Once a program has been generated, it must be executed so it can be evaluated against human created fitness tests. However, the execution of ML generated programs may include potential security risks as well as performance degradations. Because of this and the need for complex fitness tests (Section~\ref{ssect:fitness}), we developed our own interpreter. This interpreter is sandboxed within the \system\ system to provide a secure, efficient, and GA-appropriate execution environment. We explain the challenges and benefits of this system in the following subsections.

\subsubsection{Execution of Generated Programs in a Controlled Environment}

As generated programs are executed to evaluate their fitness, the results can often be undesirable and potentially dangerous. Consider a program generated with I/O instructions, allowing for the modification of files on disk. A generated program could potentially overwrite critical system files, rendering the entire machine inoperable. Likewise, a program generated with instructions to support networking could inadvertently flood a computer network (e.g., denial of service attack~\cite{zargar:2013:ddos}) or replicate itself across machines (e.g., worm~\cite{computerWorm}). 

Normally, these types of behaviors are malicious, yet ML software generators happen upon these situations in an attempt to satisfy fitness goals. By executing programs within our own secure interpreter, which includes instruction-level protection checks, \system\ can provide the additional security measures that are needed to prevent ML generated software from causing harmful behaviors. Non-ML generating interpreters and compilers do not generally include these types of checks because management of such issues are not usually within scope for HIPLs~\cite{torczon:2007:compiler}.

\subsubsection{Termination of Infinite Loops}

Automatically generated software has the potential to create infinite loops. This can occur from unsatisfied loop termination constraints or unexpected looping instructions. In our experiments, this type of behavior often arises in early program generations due to the GA maximizing the goal fitness score at the cost of program execution time. As a result, unterminated programs have the potential to halt the generation process, resulting in a failure of further program evaluations.

In an attempt to mitigate this, one can add fitness constraints to prefer programs with fewer instruction counts over larger ones. However, the generation of infinite loops, especially in early generations of programs, remains a possibility. \system's interpreter includes a customizable maximum instruction count per execution. Programs that exceed the instruction count are terminated. A fitness penalty can then be applied, reducing the likelihood of future generations of programs carrying forward the infinite loop constructs. With this addition, \system\ is guaranteed to terminate all infinite loops.

\subsubsection{Simulation of Complex Instructions}

Optimizing program execution is a principle concern for ML program generators. This is because such systems may generate and execute dozens to millions of programs before one with a high enough fitness score is found. While simple operations, such as |add|, |load|, |store|, |jmp|, may take a single clock cycle to complete, more complex operations can require many. Examples include disk I/O, networking, and peripheral device access. These types of operations can significantly increase program execution time, as they often rely on accessing services or devices with increased latency.

\system\ can simulate the execution of these complex instructions. In doing so, the GA-based programs it generates can execute more efficiently, while still retaining the ability to check the program's fitness goals. Moreover, such simulation protects the devices themselves from overuse by the plethora of programs that may attempt to access them during exploratory evolution of GA program generation.


\section{\label{sect:progamming}Using \system}

\system\ consists of a modular framework, designed in C\# .NET. It includes an engine for running genetic algorithms, an encoder and decoder for genomes, a sandboxed interpreter for simulated program execution, and a compiler to transform code into binary executables. While the initial design of \system\ uses C\#, it is important to note that the principles employed by it are not bound to C\#.

\system's software framework for fitness test construction is extensible and was developed so users can devise a myriad of customized fitness suites, which eventually guide the system's GA generation and evolution of software programs.

\paragraph{Specifying Requirements of a Program}

To generate a program, \system\ must be provided with the requirements for the desired input and output of the target program. For example, if a program should prompt a user for a numerical input and then subsequently output a line of text, this must be specified in the form of training data to AI Programmer. The following subsections detail the step-by-step process of how a program is specified and generated with AI Programmer.

\paragraph{Creating Your First Program}

To begin, a user creates a C\# class within the \system\ project, inheriting from the |FitnessBase| base class. This class includes all the necessary requirements for specifying a solution to be built by \system, including fitness scoring functionality, program termination rules, and program generation capabilities.

\paragraph{Specifying a Target Fitness Score}

Next, the user indicates a target fitness score which is specified in the constructor of the class as shown in Figure~\ref{fig:constructorFitness}. The score is typically based upon characteristics of the desired program. For example, if the target program is intended to output a string, such as ``Hello World'', the fitness score might be the number of characters in the string (i.e., 11). However, since \system\ generates programming code at the byte level, the fitness score should account for incremental differences in output characters. In this case, one should multiply each target character in the output by 256, resulting in 2816 (e.g., 11 * 256) and use that as the resulting target fitness score.

\begin{figure}[h]
\begin{VerbWithNumbers}
public StringFitness(GA ga, int maxIterCount) 
  : base(ga, maxIterCount)
{
  _targetStr = "Hello World";
  _targetFit = _targetStr.Length * 256;
}
\end{VerbWithNumbers}
\caption{Example target fitness score for ``Hello World.''}
\label{fig:constructorFitness}
\end{figure}

\system\ is designed to continue execution, generating incrementally better programs that satisfy the fitness conditions, until the current fitness score reaches its target.

\paragraph{Specifying Fitness Conditions}

Next, the user must specify the rules that are used to score each of the generated programs (i.e., the fitness test). At each generation epoch, \system\ will favor programs that have fitness scores that are closer to the target fitness score. Therefore, careful crafting of the fitness conditions are required so fitness scores accurately represent the desired goal of the program.

In the ``Hello World'' example, the user must specify that the output of the program should match the target string. To achieve this, one can add to the fitness score according to how close each character in the generated output string is to the target string. In particular, the fitness test can simply loop over the characters in the string "Hello World", and compare each one against the characters produced in the output of the generated program, adding or subtracting accordingly, as shown in Figure~\ref{fig:fitnessMethod}.

\begin{figure}[h]
\begin{VerbWithNumbers}
for (int i = 0; i < _targetStr.Length; i++)
  if (_console.Length > i)
    Fit += 256-Math.Abs(_console[i]-_targetStr[i]);
\end{VerbWithNumbers}
\caption{Adding and subtracting the fitness score based upon program output.}
\label{fig:fitnessMethod}
\end{figure}


After assigning a fitness score to each generated program within the current pool, a check is made to determine whether the target fitness score has been achieved by any of the generated programs. If so, \system\ halts and returns the solution program. Otherwise, it continues with the next generation of programs.

\paragraph{Specifying Conditions for Variable Output}

Previously, we presented an example on how to train \system\ to find an exact string. However, for more complex scenarios, such as variable outputs or calculated values, a series of training examples may be required. In such scenarios, training data can be created to serve as an initial set of examples to base the fitness score upon. Thereafter, \system\ can be guided with an evolutionary goal to generalize from the training data and provide correct results for new data.

As an example of variable output, consider the generation of a program to output the summation of two numbers. The target fitness score for this program would be the desired output, which, in this case, would be one byte, multiplied by the range of potential values (256). To construct the target fitness score, we can simply multiply the target fitness for a single result by the number of training examples. Therefore, in this case, the target fitness is |trainingCount * 256|.

After specifying the target fitness score, we can implement the actual fitness check for adding two numbers by looping over each training set combination (consisting of two numbers), inputting those values to our program, and checking the output for the correct sum. An example of this is shown in Figure~\ref{fig:addingFitness}. By providing varying training input values we can help foster the generalization of the solution program, rather than the generation of a program that only solves the exact training examples provided. 

\begin{figure}[h]
\begin{VerbWithNumbers}
int val;
if (Int32.TryParse(_console.ToString(), out val)) {
  Fit += 256 - Math.Abs(val -(input1 + input2));
}
\end{VerbWithNumbers}
\caption{Calculating the fitness of adding two numbers.}
\label{fig:addingFitness}
\end{figure}

\paragraph{Programmatic Sequences of Action}

Because different programs require different sequences of actions (e.g., requesting input, outputting a result, etc.), \system\ provides users with a mechanism to specify the necessary programmatic sequence of actions within the fitness method.

Programmatic sequences can be provided in the form of a simple state machine within the fitness check method. When the generated program executes a command to request input from the user, a bonus score can be applied to the fitness if it is executed at the correct time in the sequence of actions. Likewise, when data is output, one can add or subtract from the fitness score according to the time the action is executed. 

It is important for users to account for programmatic sequence bonuses when they are generating the initial target fitness. Doing so will ensure the generated solution will satisfy all required constraints, including sequences of events, before returning a viable solution program.

\section{\label{sect:results}Results}

Using \system, we were able to generate numerous complete software programs. A complete listing of these programs, their associated program generation time, and the total number of evolutionary generations used to build them are shown in Table~\ref{tab:bfresults}. It is important to note that the number of evolutionary generations is not equivalent to the total computational time to generate a program. This is due, in large part, to varying genome size and fitness function computation, which is unique to each program.

Even though genetic algorithms are embarrassingly parallel and \system\ utilizes task-level parallelism for each generation's genome construction and fitness test evaluation, we limited our experimental study to commodity hardware only. All experiments were run on an Intel Quad-Core i7 CPU, 2.7GHz, containing 16GB ram with an x64-based processor utilizing up to 4-threads for the parallelism described above. We constrained our experiments in this manner to demonstrate the efficacy of \system\ for real-world autonomous software development.

\begin{table}[h]
  \caption{AI Programmer Results}
  \label{tab:bfresults}
  \begin{tabular}{lllll}
    \toprule
    Name&Duration (s) &Generations\\
    \midrule
    hi & 52 & 5,700\\
    Hi! & 7,644 & 1,219,400\\
    hello & 1,713 & 252,000\\
    hello world & 7,702 & 580,900\\
    reddit & 1,362 & 195,000\\
    Keep Calm Keep Coding & 944 & 21,400\\
    I love all humans & 36,000 & 6,057,200\\
    hello \{user\} & 1,793 & 42,800\\
    Addition & 2,698 & 92,400\\
    Subtraction & 4,305 & 177,900\\
    Multiply x2 & 6,353 & 242,000\\
    Multiply x3 & 5,165 & 87,200\\
    XOR & 2,095 & 146,400\\
    Fibonacci & 21,862 & 151,900\\
    If/then conditionals & 8,313 & 46,200\\
    cats are evil & 10,209 & 814,400\\
    Bottles of Beer on the Wall & 2,957 & 61,400\\
    Reverse string & 49 & 2,600\\
    CSV parse & 173 & 9,000\\
    Extract in quotes & 6,478 & 212,100\\
    Extract in quotes 2 & 9,996 & 188,400\\
    Trim left of quote & 9,030 & 341,700\\
    XML to JSON & 6,866 & 820,900\\
    Warning countdown & 48 & 900\\
  \bottomrule
\end{tabular}
\end{table}

For the remainder of this section, we highlight the details of some of the programs listed in Table~\ref{tab:bfresults} and discuss novel aspects that emerged when generating them.

\subsection{Greetings}

``Hello World'' is usually one of the first programs human programmers create when they begin learning programming. As such, we found it fitting to guide AI Programmer to learn some basic greetings for its early programs. Rather than starting with ``Hello World'', we first had AI Programmer create a more simplistic program that simply output ``hi.'' It was successfully after 5,700 generations and the generated code is shown in Figure~\ref{fig:hi}.

\begin{figure}[h]
\begin{VerbWithNumbers}
+[+++++-+>++>++-++++++<<]>++.[+.]-.,-#>>]<]
\end{VerbWithNumbers}
\caption{Generated program: ``hi''}
\label{fig:hi}
\end{figure}

The generated program fulfilled its requirement to output the target text, but interestingly included subsequent random characters, which contained parsing errors, including non-matching brackets. However, AI Programmer's interpreter computes results until the program fails. In this manner, the syntax error (which is later on in the code, after a solution is reached) does not negatively impact its fitness score, and thus offers a working solution. In fact, the generated code can be executed in almost any third-party interpreter as a valid working program (provided, warnings are ignored).

Next, we guided AI Programmer to generate the famous ``hello world'' output which was successfully constructed after 580,900 generations and consists of the code shown in Figure~\ref{fig:helloworld}.

\begin{figure}[h]
\begin{VerbWithNumbers}
-><[>-<+++]->>++++[++++++++++++++++++<+]>.---.
+-+++++++..+++.+>+<><+[+><><>+++++++++.+-<-+++
+[++[.--------.+++.------],.-----]]
\end{VerbWithNumbers}
\caption{Generated program: ``hello world''}
\label{fig:helloworld}
\end{figure}

\paragraph{``I love all humans''}
As a humorous aside, we asked AI Programmer to create the program to output ``I love all humans,'' which was successfully generated after 6,057,200 generations. It consists of the code shown in Figure~\ref{fig:iloveallhumans}. The fitness method for this example includes a check on the output string length to ensure an exact matching output, without extraneous text.

To ensure an exact output string, the fitness score includes not just a check on the output characters, but also a check on the length of the string. In this case, the target fitness included an additional 10 points, of which a percentage of this amount is added to the resulting fitness, depending on how close the length of the output string matches the length of the target. This forces the generation of a program that outputs the exact target string, without extraneous output instructions, as the generation process will not halt until the target fitness is reached, of which, 10 points comprise having the correct output length.

\begin{figure}[h]
\begin{VerbWithNumbers}
+[>+<+++]+>------------.+<+++++++++++++++++++
++++++++++++.>+++++++++++++++++++++++++++++++
+++.+++.+++++++.-----------------.--<.>--.+++
++++++++..---<.>-.+++++++++++++.--------.----
--------.+++++++++++++.+++++.
\end{VerbWithNumbers}
\caption{Generated program: ``I love all humans''}
\label{fig:iloveallhumans}
\end{figure}

\begin{figure}[h]
\begin{VerbWithNumbers}
// Assigning the target fitness.
_targetFitness = _targetString.Length * 256;
_targetFitness += 10;

...

// Calculating the fitness length bonus.
Fitness += 10 * ((_targetString.Length -
  Math.Abs(_console.Length - 
  _targetString.Length)) /
  _targetString.Length);
\end{VerbWithNumbers}
\caption{A percentage of 10 points is added to the fitness, according to how exact the length of the output is to the target.}
\label{fig:fitnessLengthBonus}
\end{figure}

\subsection{Input-Output Computations}

We next guided AI Programmer to generate programs that perform computations based on user input. In such programs, the user provides some input and the computer program then generates the appropriate output. 

\paragraph{Reversing a String}

AI Programmer was able to generate the program to reverse any string after only 2,600 generations. The generated code is shown in Figure~\ref{fig:reversestring}. 

\begin{figure}[h]
\begin{VerbWithNumbers}
+->,>,[>+,],,,,-<[.+<]
\end{VerbWithNumbers}
\caption{Generated program for reversing a string.}
\label{fig:reversestring}
\end{figure}

\noindent
When executed, the program prompts the user for input. The user then types one character at a time until a value of ``0'' is entered. A novelty of this program is that it is required to take variable size input first before performing the majority of its program logic. However, the program's internal memory state must manage the variable input, as the program must read all input first to locate the final character entered, which is the first character in the reversed string. The genetic algorithm was able to produce this logic automatically, based upon the fitness method.

\paragraph{Addition and Subtraction}
AI Programmer was able to generate programs for addition after 92,400 generations (Figure~\ref{fig:add}) and subtraction after 177,900 generations (Figure~\ref{fig:subtract}).

\begin{figure}[h]
\begin{VerbWithNumbers}
,>,-[-<+>]<+.
\end{VerbWithNumbers}
\caption{Generated program for performing addition.}
\label{fig:add}
\end{figure}

\begin{figure}[h]
\begin{VerbWithNumbers}
,-->,-[-<->]<+.
\end{VerbWithNumbers}
\caption{Generated program for performing subtraction.}
\label{fig:subtract}
\end{figure}

\paragraph{If-Then Conditionals with User Input}

Generating programs involving more complex programming logic, such as the ability to perform if-then decisions and actions, requires a more advanced type of fitness function. However, as described in Section~\ref{sect:bfinterpret}, AI Programmer's embedded interpreter provides significantly more access to program state than just its output, which is essential for generating a large variety of more complex programs.

For example, AI Programmer was able to produce a program which prompts the user for input (e.g., 1, 2 or 3) and outputs text based on which value was entered, similar to selecting an option from a menu. By entering the value ``1'', the program would output ``hi''. Entering ``2'', resulted in the program output of ``z''. Entering ``3'', resulted in the output ``bye''. The program was generated in 446,200 generations.

The produced code was notably larger than previously generated programs, containing 650 instructions (although not all instructions are needed). The larger code was required, as the conditional branches are contained within individual blocks of the code.

\subsection{Complexity in Fitness Functions}

As the complexity of the target program grows, so too does the fitness function. After all, the fitness function needs to guide the engine in determining how well a particular child program matches the targeted solution. For conditionals and branching, successful program generation required more advanced techniques within the fitness function.

In particular, a check was needed to examine the interpreter's memory register (i.e., current data pointer via shift operations), where the distinct number of memory registers being used by the program was counted, providing a bonus to fitness to favor more memory register usage over less. This aided in inspiring diversity amongst child programs. Additionally, the instruction pointer used for each print command was recorded and weighed against the fitness score. A penalty was applied for reuse of the same print command. This helped to foster diversity and achieve a successful if-then result.

\section{Optimizing Program Generation}

We noticed that the program generation time increased significantly as the length of the target output increased. Furthermore, the need to extend AI Programmer beyond the basic instruction set was deemed a necessity if we were to have it produce programs with more interesting features, such as file I/O and networking capabilities.

As such, we extended AI Programmer to use an extended programming instruction set, which reduced code generation time and improved code compression due to an increased range of instruction specificity (i.e., fewer instructions to achieve the same result). However, a disadvantage of utilizing the extended instruction set is that the generated programs would be difficult to test in standard interpreters. As the extended instruction set for AI Programmer deviated from the traditional programming language, standard interpreters would no longer be able to run the produced code. In our case, AI Programmer's internally developed interpreter was modified to support the extended instruction set, so this was not a practical obstacle.

\subsection{Extended Instruction Set}

Several extensions of the programming language used by AI Programmer exist, which are suitable to decrease program generation time. Specifically, the speed-enhancing extension set, Extended Type III \cite{bytype3:2016:wiki}, offers several programming instructions that aid generation. These instructions include the ability to immediately set the value of a particular cell to a multiple of 16, also called ``fast cell initializers''. This aids in allowing a generated program to quickly reach displayable ASCII range characters for output, thus, decreasing the number of individual increment programming instructions that would normally be required.

In addition to key instructions taken from Extended Type III, we added several new instructions to support calling functions from within a program, allowing for increasingly complex programs to be generated.

\paragraph{Fibonacci Sequence}

With these extensions in place, AI Programmer was able to generate a program to output the Fibonacci sequence up to 233 \footnote{255 is the max value for a byte, with the next Fibonacci sequence value being 377.}, which was was generated in approximately six hours. The program prompts the user for input of the two starting values in the sequence. It then outputs the next digits in the Fibonacci sequence. The generated code for this, using the extended instruction set, is shown in Figure~\ref{fig:fibonacci}.

\paragraph{Advancing Complexity}

The ability of the GA to generate a program for solving the Fibonacci sequence was a profound advancement. The solution program contains several distinct programming tasks, including prompting the user for input of two numbers at the beginning of execution, calculating the addition of values, determining the correct mathematical sequence, outputting the result, and looping to repeat the process for each value in the sequence.

This combination of tasks, spreading across a range of programming abilities, might typically be given to human programmers in order to evaluate their programming proficiency. The capability of the GA to automatically generate this type of program demonstrates the potential for future expansion of the system.

\begin{figure}
\begin{verbatim}
,>,$[!>--$<<a>>]4]+,,-[-<+>]<+.$@
\end{verbatim}
\caption{Generated program to output the Fibonacci sequence from two starting input values.}
\label{fig:fibonacci}
\end{figure}

\section{Related Work}

Genetic programming has previously been applied in some restricted cases. A key limitation in their broader application has been in the computational density of the search space involved in program generation, which exponentially increases as programs grow in complexity~\cite{koza2010human}. \system\ provides to novel mitigation of this inefficiency by using a minimalistic programming language, exploiting the natural parallelism of GAs, simulating complex instructions, and embedding an optimized interpreter for fast execution and fast-failure of defunct programs.

\subsection{Genetic Algorithms in Other Domains}

Somewhat related to our work, is the use of program synthesis driven by genetic programming in hardware-based niche fields. Koza et al. used an automated process for creating analog circuits, involving genetically evolved designs with evolutionary computation to produce circuit components that typically require human-level intelligence to construct~\cite{koza1997automated}. In addition, they used human constructed fitness methods to guide their circuit design. Although applied in different domains, the high-level machine learning approach of Koza et al.'s system is similar to \system.

One of the key components of our research is the usage of a minimalistic programming language to limit the computational complexity of generated programs. This approach been found useful in other areas of genetic programming, including the simulation of artificial life, as described in Ling's work~\cite{ling2012artificial}. In a simulation library based on genetic algorithms and biological hierarchy, the system, called Ragaraja, uses biological concepts to form an esoteric programming language, consisting of a set of 3-character instructions. In this manner, the system is able to simplify the genetic algorithm generation and mutation process by limiting the number of possible instruction combinations. Although applied in a completely different domain, the affects of this approach are similar to \system, specifically for optimizing the generation time and limiting the complexity of generated solutions to a constrained set of instructions.

\subsection{Different Approaches in Program Generation}

\system\ has similarities to a program synthesis technique called \emph{sketching} in that each approach attempts to automatically generate software by using some human guidance. However, the similarities between the two approaches ends there. On one hand, sketching is a program synthesis technique where a programmer provides only a minimalistic outline of an implementation and the compiler generates the remaining code~\cite{armando:2005:sketch, armando:2006:sketch}. On the other hand, \system requires no partial implementation, but instead requires human developers to design fitness tests which guide the evolutionary algorithm for the entire program construction.

Another slightly related work is verified lifting~\cite{kamil:2016:lift}. Verified lifting aims to lift algorithms written in one language and place a formally verified equivalent in another language. The benefits of verified lifting are highly practical, especially when considering the need for such systems as real software systems often migrate from one programming language to another. However, verified lifting and \system\ are only loosely similar in that both systems perform automatic code generation, but do not possess any other similarities in their approaches.

\subsection{Slow Acceptance of Genetic Algorithms}

The potential capabilities of automated program generation using machine learning techniques have been considered for some time. Yet, these approaches have encountered obstacles inhibiting their practical application. Part of those obstacles were a lack of computational power and data movement throughout. Advances in these fields have had recent breakthroughs leading to the democratization of machine learning, especially in the area of deep learning, which requires complex neural networks and a large amount of training data~\cite{goodfellow:2017:dl}.

Still, other challenges remain in automated programming, as explained by O'Neil et al.~\cite{o2010open}, which describes the slow growth of genetic programming, despite the successes that it has achieved in various real-world domains. To the best of our knowledge, \system\ is the first end-to-end GA system to demonstrate rapid progress in non-trivial program generation achieved entirely on commodity hardware.

\section{Conclusion and Future Work}
Traditional human-based computer programming is approaching a dramatic shift. With increasingly complex software and hardware advances and the growing challenges integrating the two, the craft of software development will inevitably surpass the capabilities of humans. As that time approaches, it will be necessary to have some form of automatic software generation to assist humans in software development beyond what exists today (e.g., compilers, higher-level programming languages, etc.).

The results presented in this paper, provide early notions about the power that machine learning techniques, specifically genetic algorithms, may offer a partial solution for automatic program generation. We showed that fully functional programs can indeed be automatically generated, provided they are supplied with some human guidance in the way of input parameters and training data. While the initial set of programs generated by \system\ are similar in complexity to that a novice human programmer, the range of generated programs need not restricted to traditional means such as human time or human intellect. Instead, they are simply a function of fitness test complexity and computational resources.

In addition to correctness, efficient implementation of fitness methods are imperative to the practical application of \system. This is because each generated program is checked against the fitness method every time a new program is evaluated. An important open area of future work is the deep examination of how to implement fitness methods as efficiently as possible while still retaining a high degree of correctness. One possible solution is to build superoptimizers specifically for fitness test optimizations~\cite{phothilimthana:2016:superopti}.

Another important open area in ML-based program generation is the need for specifically crafted programming languages that have strong alignment with the constraints of ML computation. The current programming languages we use today, for humans, are ill-suited for ML-based program generation. The approach we use for typical program language creation needs to be abandoned and rethought when considering a future of ML-driven program generation. Only once this is done, can we begin to envision a new future of computer software development, driven by artificial intelligence based systems, with human creativity and design guiding the way.



%


\bibliographystyle{abbrvnat}
\bibliography{ml}





\end{document}